%% file: main.tex
\documentclass[10pt,twocolumn,letterpaper]{article}

\usepackage{cvpr}
\usepackage{times}
\usepackage{epsfig}
\usepackage{graphicx}
\usepackage{amsmath}
\usepackage{amssymb}

\usepackage{epstopdf}
\usepackage{enumitem}
\usepackage{multirow}
\usepackage[]{algorithm2e}

\input{math_definition}

\usepackage[pagebackref=true,breaklinks=true,letterpaper=true,colorlinks,bookmarks=false]{hyperref}

\cvprfinalcopy

\ifcvprfinal\fi
\begin{document}

\title{Cross-Dataset Adaptation for Visual Question Answering}

\author{
Wei-Lun Chao$^*$\\
U. of Southern California\\
Los Angeles, CA\\
{\tt\small weilunchao760414@gmail.com}
\and
Hexiang Hu\thanks{\hspace{4pt}Equal contributions}\\
U. of Southern California\\
Los Angeles, CA\\
{\tt\small hexiang.frank.hu@gmail.com}
\and
Fei Sha\\
U. of Southern California\\
Los Angeles, CA\\
{\tt\small feisha@usc.edu}
}

\maketitle

\input{abs}

\input{intro}
\input{related}
\input{bias}
\input{method}
\input{exp}

\input{disc}

\input{ack}

{\small
\bibliographystyle{ieee}
\bibliography{vqa}
}

\clearpage
\appendix
\begin{center}
	\textbf{\Large Supplementary Material}
\end{center}

We provide contents omitted in the main text.
\begin{itemize}
	\item Section~\ref{s_NTD}: details on \emph{Name that dataset!} (Sect.~3.2 of the main text).
	\item Section~\ref{s_Alg}: details on the proposed domain adaptation algorithm (Sect.~4.2 and 4.3 of the main text).
	\item Section~\ref{s_set}: details on the experimental setup (Sect. 5.2 of the main text).
	\item Section~\ref{s_exp}: additional experimental results (Sect. 5.3 and 5.4 of the main text).
\end{itemize}

\input{suppl_algorithms}
\input{suppl_results}

\end{document}

%% file: math_definition.tex
\usepackage{amsmath,amsfonts}
\usepackage{amssymb,amsopn}
\usepackage{bm} 

\usepackage{amssymb}
\usepackage{amsmath,amsfonts}
\usepackage{amsthm,amsopn}

\usepackage{bm} 

\newcommand{\vct}[1]{\boldsymbol{#1}} 
\newcommand{\cst}[1]{\mathsf{#1}}  

\newcommand{\ProbOpr}[1]{\mathbb{#1}}

\newcommand{\expect}[2]{%
\ifthenelse{\equal{#2}{}}{\ProbOpr{E}_{#1}}
{\ifthenelse{\equal{#1}{}}{\ProbOpr{E}\left[#2\right]}{\ProbOpr{E}_{#1}\left[#2\right]}}} 
\newcommand{\var}[2]{%
\ifthenelse{\equal{#2}{}}{\ProbOpr{VAR}_{#1}}
{\ifthenelse{\equal{#1}{}}{\ProbOpr{VAR}\left[#2\right]}{\ProbOpr{VAR}_{#1}\left[#2\right]}}} 

\DeclareMathOperator{\argmax}{arg\,max}
\DeclareMathOperator{\argmin}{arg\,min}

\newcommand{\vtheta}{\vct{\theta}}

\newcommand{\vphi}{\vct{\phi}}

\newcommand{\eat}[1]{}

%% file: abs.tex
\begin{abstract}

We investigate the problem of cross-dataset adaptation for visual question answering (Visual QA). Our goal is to train a Visual QA model on a source dataset  but apply it to another target one. Analogous to domain adaptation for visual recognition, this setting is appealing when the target dataset does not have a sufficient amount of labeled data to learn an ``in-domain'' model.  The key challenge is that the two datasets are constructed differently, resulting in the cross-dataset mismatch on images, questions, or answers. 

We overcome this difficulty by proposing a novel domain adaptation algorithm. Our method reduces the difference in statistical distributions by transforming the feature representation of the data in the target dataset. Moreover, it maximizes the likelihood of answering questions (in the target dataset) correctly using the Visual QA model trained on the source dataset. We empirically studied the effectiveness of the proposed approach on adapting among several popular Visual QA datasets. We show that the proposed method improves over baselines where there is no adaptation and several other adaptation methods. We both quantitatively and qualitatively analyze when the adaptation can be mostly effective. 
\end{abstract}

%% file: intro.tex
\section{Introduction}
Visual Question Answering (Visual QA) has emerged as a very useful task to pry into how well learning machines can comprehend and reason with both visual and textual information, which is an important functionality for general artificial intelligence. In this task, the machine is  presented with an image and a relevant question. The machine can generate a free-form answer or select from a pool of candidates.  In the last few years, more than a dozen datasets for the task have been developed~\cite{gupta2017survey, kafle2016visual,wu2016visual}. Despite a steady and significant improvement in modeling, the gap in performance between humans and machines on those datasets is still substantial. For instance, on the VQA dataset~\cite{antol2015vqa} where human attains accuracy of 88.5\%, the state-of-the-art model on the Visual QA multiple-choice task achieves 71.4\%~\cite{yu2017multi}.

In this paper, we study another form of performance gap. Specifically, \emph{can the machine learn knowledge well enough on one dataset so as to answer adeptly questions from another dataset}?  Such study will highlight the similarity and difference among different datasets and guides the development of future ones. It also sheds lights on how well learning machines can understand visual and textual information in their generality, instead of learning and reasoning with dataset-specific knowledge.

\begin{figure}
\centering
\includegraphics[width=0.49\textwidth]{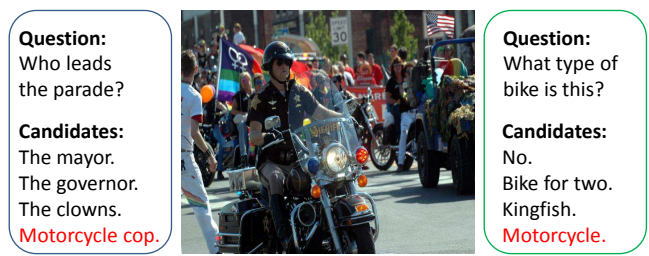}
\caption{An illustration of the dataset bias in visual question answering. Given the same image, Visual QA datasets like VQA~\cite{antol2015vqa} (right) and Visual7W~\cite{zhu2016visual7w} (left) provide different styles of questions, correct answers (red), and candidate answer sets, each can contributes to the bias to prevent cross-dataset generalization.}
\label{f_Concept}
\vskip -1.5em
\end{figure}

Studying the performance gap across datasets is reminiscent of the seminal work  by Torralba and Efros~\cite{torralba2011unbiased}. There, the authors study the bias in image datasets for object recognition. They have showed that the idiosyncrasies in the data collection process cause domain mismatch such that classifiers learnt on one dataset degrade significantly on another dataset~\cite{gong2012geodesic,gong2013connecting,gong2013reshaping,khosla2012undoing,mclaughlin2015data,tommasi2014testbed,Herranz_2016_CVPR,tommasi2015deeper}. 

The language data in the Visual QA datasets introduces an addition layer of difficulty to bias in the visual data (see Fig.~\ref{f_Concept}). For instance, ~\cite{ferraro2015survey} analyzes several datasets and illustrates their difference in syntactic complexity as well as within- and cross-dataset perplexity. As such, data in Visual QA datasets are likely more taletelling the origins from which datasets they come. 

To validate this hypothesis, we had designed a \emph{Name That Dataset!} experiment, similar to the one in ~\cite{torralba2011unbiased} for comparing visual object images. We show that the two popular Visual QA datasets VQA~\cite{antol2015vqa} and Visual7W~\cite{zhu2016visual7w} are almost complete distinguishable using either the question or answer data. See Sect.~\ref{S_bias} for the details of this experiment.

Thus, Visual QA systems that are optimized on one of those datasets can focus on dataset-specific knowledge such as the type of questions as well as how the questions and answers are phrased. This type of bias exploitation hinders cross-dataset generalization and does not result in AI systems that can reason well over vision and text information in different or new characteristics.

In this paper, we investigate the issue of cross-dataset generalization in  Visual QA. We assume that there is a source domain with a sufficiently large amount of annotated data such that a strong Visual QA model can be built, albeit adapted to the characteristics of the source domain well. However, we are interested in using the learned system to answer questions from another (target) domain. The target domain does not provide enough data to train a Visual QA system from scratch.  We show that in this domain-mismatch setting, applying directly the learned system from the source to the target domain results in poor performance.

We thus propose a novel adaptation algorithm for Visual QA. Our method has two components. The first is to reduce the difference in statistical distributions by transforming the feature representation of the data in the target dataset. We use an adversarial type of loss to measure the degree of differences---the transformation is optimized such that it is difficult to detect the origins of the transformed features.

The second component is to maximize the likelihood of answering questions (in the target dataset) correctly using the Visual QA model trained on the source dataset. This ensures the learned transformation from optimizing domain matches retaining the semantic understanding encoded in the Visual QA model learned on the source domain.

The rest of this paper is organized as follows. In Sect.~\ref{S_related}, we review related work. In Sect.~\ref{S_bias}, we analyze the dataset bias via the game \emph{Name That Dataset!} In Sect.~\ref{DA_task}, we define tasks of domain adaptation for Visual QA. In Sect.~\ref{S_method}, we describe the proposed domain adaptation algorithm. In Sect.~\ref{S_exp}, we conduct extensive experimental studies and further analysis. Sect.~\ref{S_disc} concludes the paper.

%% file: related.tex
\section{Related Work}
\label{S_related}
\paragraph{Datasets for Visual QA} 

About a dozen of Visual QA datasets have been created~\cite{kafle2016visual,wu2016visual,gupta2017survey,goyal2016making, kafle2017analysis,agrawal2018don}. In all the datasets,  there are a collection of \textbf{images (I)}.  Most of existing datasets use natural images from large-scale common image databases (e.g. MSCOCO~\cite{lin2014mscoco}). For each image, human annotators are asked to generate multiple \textbf{questions (Q)} and to provide the corresponding \textbf{``correct"  answers (T)}. This gives rise to image-question-correct answer (IQT) triplets.  Visual7W~\cite{zhu2016visual7w} and VQA~\cite{antol2015vqa} further include artificially generated \textbf{``negative" candidate answers (D)}, referred as decoys, for the multiple-choice setting. Dataset biases can occur in any of these steps in creating the datasets. 

\vspace{-10pt}
\paragraph{Tasks} While the machine can generate free-form answers, evaluating the answers is challenging and not amenable to automatic evaluation. Thus, so far a convenient paradigm is to evaluate machine systems using multiple-choice based Visual QA~\cite{antol2015vqa,chao2017being,zhu2016visual7w,jabri2016revisiting}. The machine is presented the correct answer, along with several decoys (incorrect ones) and the aim is to select the right one.  The evaluation is then automatic: one just needs to record the accuracy of selecting the right answer. Alternatively, the other setting is to select one from the top frequent answers and compare it to multiple human-annotated ones~\cite{anderson2017bottom,antol2015vqa,ben2017mutan,fukui2016multimodal,goyal2016making,kafle2017analysis,lu2016hierarchical,xu2016ask,yang2016stacked,yu2017multilevel,yu2017multi}, avoiding constructing decoys that are too easy such that the performance is artificially boosted~\cite{antol2015vqa,goyal2016making}.

\vspace{-10pt}
\paragraph{Methods for Visual QA} As summarized in~\cite{kafle2016visual,wu2016visual,gupta2017survey}, one popular framework of Visual QA algorithms is to learn a joint image-question embedding, e.g., by the attention mechanism, followed by a multi-way classifier (among the top frequent answers)~\cite{yu2017multi, anderson2017bottom, ben2017mutan,fukui2016multimodal, yang2016stacked, lu2016hierarchical}. Though lacking the ability to generate novel answers beyond the training set, this framework has been shown to outperform those who can truly generate free-form answers. For the multiple-choice setting, one line of algorithms is to learn a scoring function with image, question, and a candidate answer as the input. Even a simple multi-layer perceptron (MLP) model achieves the state of the art~\cite{jabri2016revisiting,fukui2016multimodal,shih2016look}.

\vspace{-10pt}
\paragraph{Bias in vision and language datasets}
In~\cite{ferraro2015survey}, Ferraro et al. surveyed several exiting image captioning and Visual QA datasets in terms of their linguistic patterns. They proposed several metrics including perplexity, part of speech distribution, and syntactic complexity to characterize those datasets, demonstrating the existence of the reporting bias---the frequency that annotators write about actions, events, or states does not reflect the real-world frequencies. However, they do not explicitly show how such a bias affects the downstream tasks (i.e., Visual QA and captioning). 

Specifically for Visual QA, there have been several work discussing the bias \emph{within a single dataset}~\cite{goyal2016making,zhang2016yin,jabri2016revisiting,johnson2016clevr,chao2017being}. For example, \cite{goyal2016making,zhang2016yin} argue the existence of priors on answers given the question types and the correlation between the questions and answers (without images) in VQA~\cite{antol2015vqa}, while \cite{chao2017being} points out the existence of bias in creating decoys. They propose to augment the original datasets with additional IQT triplets or decoys to resolve such issues. \cite{jabri2016revisiting,agrawal2018don} studies biases across datasets, and show the difficulties in transferring learned knowledge across datasets. 

Our work investigates the causes of the poor cross-dataset generalization and proposes to resolve them via domain adaptation. Those causes are orthogonal to biases in a single dataset that a learning model can exploit. Thus, merely improving the Visual QA model's performance on in-domain datasets does not imply reducing the cross-dataset generalization gap, as shown in Sect.~\ref{S_bias}.

\vspace{-10pt}
\paragraph{Domain adaptation (DA)}
Extensive prior work has been done to adapt the domain mismatch between datasets~\cite{tzeng2015simultaneous,ganin2016domain,tzeng2017adversarial,chen2017show,gong2016domain,gong2013reshaping}, mostly for visual recognition while we study a new task of Visual QA. One popular method is to learn a transformation that aligns source and target domains according to a certain criterion. Inspired by the recent flourish of Generative Adversarial Network~\cite{goodfellow2014generative}, many algorithms~\cite{ganin2016domain,tzeng2017adversarial,chen2017show, yang2017semi} train a domain discriminator as a new criterion for learning such a transformation.  Our method applies a similar approach, but aims to perform adaptation simultaneously on data with multiple modalities (i.e., images, questions, and answers). To this end, we leverage the Visual QA knowledge learned from the source domain to ensure that the transformed features are semantically aligned. Moreover, in contrast to most existing methods, we learn the transformation from the target domain to the source one, similar to \cite{sun2016return, tzeng2017adversarial}\footnote{Most DA algorithms, when given a target domain, adjust the features for both domains and retrain the source model on the adjusted features---they need to retrain the model when facing a new target domain. Note that \cite{sun2016return, tzeng2017adversarial} do not incorporate the learned source-domain knowledge as ours.}, enabling applying the learned Visual QA model from the source domain without re-training. 

%% file: bias.tex
\section{Visual QA and Bias in the Datasets}
\label{S_bias}
\begin{table*}[t]
\centering
\small
\caption{Results of \emph{Name That Dataset!} }
\vspace{5pt}
\begin{tabular}{l|c|c|c|c|c|c|c|c|c}\hline
Information  & I & Q & T & D & Q + T & Q + D & T + D& Q + T + D & Random \\
\hline
Accuracy & 52.3\% & 76.3\% & 74.7\% & 95.8\% & 79.8\% &  97.5\% &  97.4\% & 97.5\% & 50.00\% \\\hline
\end{tabular}
\vskip -10pt
\label{exp:bias}
\end{table*}

In what follows, we describe a simple experiment \emph{Name That Dataset!} to illustrate  the biases in Visual QA datasets ---questions and answers are idiosyncratically constructed such that a classifier can easily tell one apart from the other by using them as inputs. We then discuss how those biases give rise to poor cross-dataset generalization errors.

\subsection{Visual QA}
\label{ss_vqa}
In Visual QA datasets, a training or test example is a IQT triplet that consists of an image I, a question Q, and a (ground-truth) correct answer T\footnote{Some datasets provide multiple correct answers to accommodate the ambiguity in the answers.}. During evaluation or testing, given a pair of I and Q, a machine needs to \emph{generate} an answer that matches exactly or is semantically similar to T. 

In this work, we focus on multiple-choice based Visual QA, \emph{since the two most-widely studied datasets---VQA~\cite{antol2015vqa} and Visual7W~\cite{zhu2016visual7w}---both consider such a setting}. In this setting, the correct answer T is accompanied by a set of $K$ ``negative" candidate answers, resulting in a candidate answer set A consist of a single T and $K$ decoys denoted by D. An IQA triplet is thus $\{\text{I}, \text{Q},\text{A}=\{\text{T},\text{D}_1,\cdots,\text{D}_K\}\}$. We use C to denote an element in A. During testing, given I, Q, and A, a machine needs to select T from A. Multiple-choice based Visual QA has the benefit of simplified evaluation procedure and has been popularly studied~\cite{jabri2016revisiting,yu2017multilevel,fukui2016multimodal,shih2016look,kembhavi2016diagram}. Note that in the recent datasets like VQA2~\cite{goyal2016making}, the candidate set A is expanded to include the most frequent answers from the whole training set, instead of a smaller subset typically used in earlier datasets. Despite this subtle difference, we do not lose in generality by studying cross-dataset generalization with multiple-choice based Visual QA datasets.

\begin{figure}[t]
\centering
\includegraphics[width=0.35\textwidth]{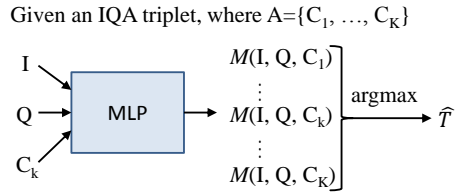}
\caption{An illustration of the MLP-based model for multiple-choice Visual QA. Given an IQA triplet, we compute the $M(I,Q,C_k)$ score for each candidate answer $C_k$. The candidate answer that has the highest score is selected as the model's answer.} 
\label{f_MLP}
\vskip -1.5em
\end{figure}

We follow~\cite{jabri2016revisiting} to train one-hidden-layer MLP models for multiple-choice based Visual QA. The MLP $M$ takes the concatenated features of an IQC triplet as input and outputs a compatible score $M$(I, Q, C) $\in[0,1]$, measuring how likely C is the correct answer to the IQ pair. During training, $M$ is learned to maximize the binary cross-entropy, where each IQC triplet is labeled with 1 if C is the correct answer; 0, otherwise. During testing, given an IQA triplet, the C $\in$ A that leads to the highest score is selected as the model's answer. We use the penultimate layer of ResNet-200~\cite{he2016deep} as visual features to represent I and the average \textsc{word2vec} embeddings~\cite{mikolov2013distributed} as text features to represent Q and C, as in~\cite{jabri2016revisiting}. See Fig.~\ref{f_MLP} for an illustration.

\subsection{Bias in the Datasets}
\label{S_bias_NTD}
We refer the term ``bias'' to any idiosyncrasies in the datasets that learning algorithms can overfit to and cause poor cross-dataset generalization.

\vspace{-10pt}
\paragraph{\emph{Name That Dataset!}} To investigate the degree and the cause of the bias, we construct a game \emph{Name That Dataset!}, similar to the one described in \cite{torralba2011unbiased} for object recognition datasets. In this game,  the machine has access to the examples (ie, either IQT or IQA triplet) and needs to decide which dataset those examples belong to. We experiment on two popular datasets Visual7W~\cite{zhu2016visual7w} and VQA~\cite{antol2015vqa}. We use the same visual and text features described in Sect.~\ref{ss_vqa} to represent I, Q, T, and D\footnote{Visual7W~\cite{zhu2016visual7w} has 3 decoys per triplet and VQA~\cite{antol2015vqa} has 17 decoys. For fair comparison, we subsample 3 decoys for VQA. We then average the \textsc{word2vec} embedding of each decoy to be the feature of decoys.}. We then concatenate these features to form the joint feature. We examine different combination of I, Q, T, D as the input to a one-hidden-layer MLP for predicting the dataset from which the sample comes. We sample 40,000, 5,000 and 20,000 triplets from each dataset and merge them to be the training, validation and test sets. Details are in the Suppl.

As shown in Table~\ref{exp:bias}, all components but images lead to strong detection of the data origin, with the decoys contributing the most (i.e., 95.8\% alone). Combining multiple components further improve the detection accuracy, suggesting that  datasets contain different correlations or relationships among components. Concatenating all the components together results in nearly 100\% classification accuracy. In other words, the image, question, and answers in each dataset are constructed characteristically.  Their distributions (in the joint space) are sufficiently distant from each other.  Thus, one would not expect a Visual QA system trained on one dataset to work well on the other datasets. See below for results validating this observation.

\vspace{-10pt}
\paragraph{Question Type is just one biasing factor} Question type is an obvious culprit of the bias. In Visual7W, questions are mostly  in the 6$W$ categories (ie, what, where, how, when, why, who). On the other hand, the VQA dataset  contains additional questions whose correct answers are either Yes or No. Those questions  barely start  with the 6$W$ words.  We create a new dataset called VQA$^-$ by removing the Yes or No questions from the original VQA dataset. 

We reran the \emph{Name That Dataset!} (after retraining on the new dataset). The accuracies of using Q or Q+T have dropped 
from 76.3\% and 79.8\% to 69.7\% and 73.8\%, respectively, which are still noticeably higher than 50\% by chance. This indicates that the questions or correct answers may phrased differently between the two datasets (e.g., the length or the use of vocabularies). Combining them with the decoys (i.e., Q+T+D) raises the accuracy to 96.9\%, again nearly distinguishing the two datasets completely. This reflects that the incorrect answers must be created very differently across the two datasets (In most cases, decoys are freely selected by the data collectors---being incorrect answers to the questions affords the data collectors to sample from unconstrained spaces of possible words and phrases.)

\vspace{-10pt}
\paragraph{Poor cross-dataset generalization}
\label{S_bias_CDG}
Using the model described in Sect.~\ref{ss_vqa}, we obtain the Visual QA accuracies of 65.7\% and 55.6\% on Visual7W and VQA$^-$ when training and testing using the same dataset. However, when the learned models are applied to the other dataset, the performance drops significantly to 53.4\% (trained on VQA$^-$ but applied to Visual7W) and 28.1\% (trained on  Visual7W but applied to VQA$^-$). See Table~\ref{t_dataset_1} for the details.

We further evaluate a variant of the spatial memory network~\cite{xu2016ask}, a more sophisticated Visual QA model. A similar performance drop is observed. See Table~\ref{t_attention} for details.

%% file: method.tex
\section{Cross-Dataset Adaptation}
\label{DA_task}

We propose to overcome the cross-dataset bias (and the poor cross-dataset generalization) with the idea of \emph{domain adaptation}. Similar ideas have been developed in the past to overcome the  dataset bias for object recognition~\cite{saenko2010adapting,gong2012geodesic}.  

\subsection{Main Idea}

We assume that we have a source domain (or dataset) with plenty of annotated data in the form of Image-Question-Candidate Answers (IQA) triplets such that we can build a strong Visual QA system. We are then interested in applying this system to the target domain. However, we do not assume there is any annotated data (i.e., IQA/IQT triplets) from the target domain such that re-training (either using the target domain alone or jointly with the source domain) or fine-tuning~\cite{oquab2014learning,wiese2017neural} the system is feasible\footnote{Annotated data from the target data, if any, can be easily incorporated into our method as a supervised learning discriminative loss. We leave this for a full version of the current paper.}.

Instead, the target domain provides \emph{unsupervised data}. The target domain could provide images,  images and questions (without either correct or incorrect answers),  questions, questions with either correct or incorrect answers or both, or simply a set of candidate answers (either correct or incorrect or both).  This last two scenarios are particularly interesting\footnote{Most existing datasets are derived from MSCOCO. Thus there are limited discrepancies between images, as shown in the column I in Table~\ref{exp:bias}. Our method can also be extended to handle large discrepancy in images. Alternatively, existing methods of domain adaptation for visual recognition could be applied to images first to reduce the discrepancy.}. From the results in Table~\ref{exp:bias}, the discrepancy in textual information is a major contributor to domain mismatch, cf. the columns starting Q.

Given the target domain data, it is not feasible to train an ``in-domain'' model with the data (as it is incomplete and unsupervised). We thus need to model jointly the source domain supervised data and the target domain data that reflect distribution mismatch. Table~\ref{t_dataset} lists the settings we work on.

\begin{table}[t]
\small
\tabcolsep 5pt
\centering
\caption{Various Settings for cross-dataset Adaptation. Source domain always provide I, Q and A (T+D) while the target domain provides the same \emph{only} during testing.}
\label{t_dataset}
\vspace{5pt}
\begin{tabular}{|l|l|} \hline
Shorthand & Data from Target at Training\\
\hline
Setting[Q] &  Q \\
Setting[Q+T] (or [Q+T+D]) & Q, T (or {Q, T+D})\\  
Setting[T] (or [T+D])  & T (or T+D) \\
\hline
\end{tabular}
\vskip -10pt
\end{table}

\subsection{Approach}
\label{S_method}

Our approach has two components. In the first part, we match features encoding questions and/or answers across two domains. In the second part, we ensure the correct answers from the target domain have higher likelihood in the Visual QA model trained on the source domain.  Note that we do not re-train the Visual QA model as we do not have access to complete data on the target domain.

\vspace{-10pt}
\paragraph{Matching domain}  The main idea is to transform features computed on the target domain (\textsf{TD}) to match those features computed on the source domain (\textsf{SD}). To this end, let $g_q(\cdot)$  and $g_a(\cdot)$ denote the transformation for the features on the questions and on the answers respectively.  We also use $f_q$, $f_t$, $f_d$, and $f_c$ to denote feature representations of a question, a correct answer, an incorrect decoy, or a candidate answer. In the Visual QA model, all these features are computed by the average \textsc{word2vec} embeddings of words.

The matching is computed as the Jensen-Shannon Divergence (JSD) between the two empirical distributions across the datasets. For the Setting[Q], the matching is
\begin{equation}
m(\textsf{TD}\rightarrow \textsf{SD}) = JSD( \hat{p}_{\textsf{SD}}(f_q), \hat{p}_{\textsf{TD}}(g_q(f_q)))
\end{equation}
where $\hat{p}_{\textsf{SD}}(f_q)$ is the empirical distribution of the questions in the source domain and $\hat{p}_{\textsf{TD}}(g_q(f_q)))$ is the empirical distribution of the questions in the target domain, after being transformed with $g_q(\cdot)$,

The JSD divergence between two distributions $P$ and $P'$ is computed as 
\begin{equation}
\medmuskip=0mu
\thinmuskip=0mu
\thickmuskip=0mu
JSD(P,P') = \frac{1}{2}\left\{KL\left(P; \frac{P+P'}{2}\right) + KL\left(P'; \frac{P+P'}{2}\right)\right\}, 
\end{equation}
while $KL$ is the KL divergence between two distributions. The JSD divergence is closely related to discriminating two distributions with a binary classifier~\cite{goodfellow2014generative} but difficult to compute. We thus use an adversarial lose to approximate it. See the Suppl. for details.

For both the Setting[Q+T] and the Setting[Q+T+D], the matching is
\begin{equation}
m(\textsf{TD}\rightarrow \textsf{SD}) = JSD( \hat{p}_{\textsf{SD}}(f_q, f_t), \hat{p}_{\textsf{TD}}(g_q(f_q), g_a(f_t)))
\end{equation}
with the empirical distributions computed over both the questions and the correct answers. Note that even when the decoy information is available, we deliberately ignore them in computing domain mismatch. This is because the decoys can be designed very differently even for the same IQT triplet. Matching the distributions of D thus can cause undesired mismatch of T since they share the same transform during testing\footnote{Consider the following highly contrived example. To answer the question ``what is in the cup?'', the annotators in the source domain could answer with ``water'' as the correct answer, and ``coffee'', ``juice'' as decoys, while the annotators in the target domain could answer with ``sparkling water'' (as that is the correct answer), then ``cat'' (as in cupcats), and ``cake'' (as in cupcakes) as decoys. While it is intuitive to match the distribution of correct answers, it makes less sense to match the distributions of the decoys as they are much more dispersed.}.

For the Setting[T] and Setting[T+D], the matching is
\begin{equation}
m(\textsf{TD}\rightarrow \textsf{SD}) = JSD( \hat{p}_{\textsf{SD}}(f_t), \hat{p}_{\textsf{TD}}(g_a(f_t)))
\end{equation}
while the empirical distributions are computed over the correct answers only.

\vspace{-10pt}
\paragraph{Leverage Source Domain for Discriminative Learning}

In the Setting[Q+T], Setting[Q+T+D], Setting[T] and Setting[T+D], the learner has access to the correct answers T (and the incorrect answers D) from the target domain. As we intend to use the transformed feature $g_q(f_q)$ and $g_a(f_c)$ with the Visual QA model trained on the source domain, we would like those transformed features to have high likelihood of being correct (or incorrect).

To this end, we can leverage the source domain's data which always contain both T and D. The main idea is to construct a Visual QA model on the source domain using the same partial  information as in the target domain, then to assess how likely the transformed features remain to be correct (or incorrect).

In the following, we use the Setting[Q+T+D] as an example (other settings can be formulated similarly). Let $h_{\textsf{SD}}(q, c)$ be a model trained on the source domain such that it  tells us the likelihood an answer $c$ can be correct with respect to  question $q$. Without loss of generality, we assume $h_{\textsf{SD}}(q, c)$ is the output of a binary logistic regression.

To use this model on the target data, we compute the following loss for every pair of question and candidate answer:
\[
\ell(q, c)  = \left\{\begin{array}{ll}
- \log h_{\textsf{SD}}(g_q(f_q), q_a(f_c)) & \hspace{-5pt}\text{if $c$ is correct,}\\
- \log (1- h_{\textsf{SD}}(g_q(f_q), q_a(f_c))) & \hspace{-5pt}\text{otherwise.}
\end{array}\right.
\]
The intuition is to raise the likelihood of any correct answers and lowering the likelihood of any incorrect ones.   Thus, even we do not have a complete data for training models on the target domain discriminatively, we have found a surrogate to minimize,
\vskip -5pt
\begin{equation}
\hat{\ell}_\textsf{TD} = \sum_{(q, c) \in \textsf{TD}} \ell(q, c),
\end{equation}
measuring all the data provided in the target data and how they are likely to be correct or incorrect. 

\subsection{Joint optimization}

We learn the feature transformation by jointly balancing the domain matching and the discriminative loss surrogate
\begin{align}
\argmin_{g_q, g_a} m(\textsf{TD}\rightarrow \textsf{SD}) + \lambda \hat{\ell}_\textsf{TD} \label{DA_obj}.
\end{align}
We select $\lambda$ to be large while still allowing $m(\textsf{TD}\rightarrow \textsf{SD})$ to decrease in optimization: $\lambda$ is 0.5 for Setting[Q+T+D] and Setting[T+D], and 0.1 for the other experiments. 
The learning objective can be similarly constructed when the target domain provides Q and T, T, or T+D, as explained above. If the target domain only provides Q, we omit the term $\hat{\ell}_\textsf{TD}$.

Once the feature transformations are learnt, we use the Visual QA model on the source domain $M_{\textsf{SD}}$, trained using image, question, and answers all together to make an inference on an IQA triplet $(i,q, A)$ from the target
\[
\hat{t} = \argmax_{c\in A} M_{\textsf{SD}}(f_i, g_q(f_q), g_a(f_c)),
\]
where we identify the best candidate answer from the pool of the correct answers and their decoys $A$ using the source domain's model. See Sect.~\ref{S_exp_setup} and the Suppl. for the parameterization of $g_q(\cdot)$ and $g_a(\cdot)$, and details of the algorithm.

%% file: exp.tex
\section{Experiments}
\label{S_exp}

\subsection{Dataset}
\input{exp_dataset}

\subsection{Experimental setup}
\label{S_exp_setup}
\input{exp_setup}

\subsection{Experimental Results on Visual7W and VQA$-$}
\input{exp_results}

\subsection{Experimental Results across five datasets}
\input{exp_across}

%% file: exp_dataset.tex
We first evaluate our algorithms on the domain adaptation settings defined in Sect.~\ref{DA_task} between Visual7W~\cite{zhu2016visual7w} and VQA~\cite{antol2015vqa}. Experiments are conducted  on both the original datasets and a revised version~\cite{chao2017being} of them. We then include Visual Genome~\cite{krishna2016vg} with the decoys created by~\cite{chao2017being} and apply the same procedure to create decoys for the COCOQA~\cite{ren2015exploring} and VQA2~\cite{goyal2016making} datasets, leading to a comprehensive study of cross-dataset generalization. 

\vspace{-10pt}
\paragraph{VQA Multiple Choice~\cite{antol2015vqa}} The dataset uses images from the MSCOCO~\cite{lin2014mscoco} dataset, with the same split setting. It contains 248,349/121,512/244,302 IQA triplets for training/validation/test. Each triplet has 17 decoys, where in general 3 decoys are human-generated, 4 are randomly sampled, and 10 are from fixed set of high frequency answers.

\vspace{-10pt}
\paragraph{Visual7W Telling~\cite{zhu2016visual7w}} The dataset contains 47,300 images from MSCOCO~\cite{lin2014mscoco} and in total 139,868 IQA triplets (69,817/28,020/42,031 for training/validation/test). Each triplet has 3 decoys: all of them are human-generated.

\vspace{-10pt}
\paragraph{Curated Visual7W \& VQA~\cite{chao2017being}} These datasets are revised versions of Visual7w and VQA, in which the decoys are carefully designed to prevent machines from ignoring the visual information, the question, or both while still doing well on the task. Each IQT triplet in Visual7W and VQA are augmented with 6 auto-generated decoys as candidate answers. Since \cite{chao2017being} only provide revised decoys for the training and validation splits, for all the studies on VQA we report results on the validation set.

\vspace{-10pt}
\paragraph{Visual Genome~\cite{krishna2016vg} \& COCOQA~\cite{ren2015exploring} \& VQA2~\cite{goyal2016making}}
These three datasets only provide IQT triplets, while~\cite{chao2017being} creates decoys for Visual Genome (VG). We required the codes from the authors of~\cite{chao2017being}, and apply the same procedure to create decoys for COCOQA and VQA2---each IQT triplet is augmented with 6 auto-generated decoys. The resulting datasets have 727,751/283,666/433,905 IQA triplets for training/validation/test on VG, 78,736/38,948 for training/test on COCOQA, and 443,757/214,354 for training/validation on VQA2. All datasets use images from MSCOCO~\cite{lin2014mscoco}.

\vspace{-10pt}
\paragraph{Evaluation metric}
For Visual7W, VG, and COCOQA, we compute the accuracy of picking the correct answer from multiple choices. For VQA and VQA2, we follow its protocol to compute accuracy, comparing the picked answer to the 10 human-annotated correct answers. The accuracy is computed based on the number of exact matches among the 10 answers (divided by 3 and clipped at 1).

%% file: exp_setup.tex
\paragraph{Visual QA model}
In all our experiments, we use a one hidden-layer MLP model (with 8,192 hidden nodes and ReLU) to perform binary classification on each input IQC (image, question, candidate answer) triplet, following the setup as in \cite{jabri2016revisiting,chao2017being}. Please see Fig.~\ref{f_MLP} and Sect.~\ref{ss_vqa} for explanation. The candidate C $\in$ A that has the largest score is then selected as the answer of the model. Such a simple model has achieved the state-of-the-art results on Visual7W and comparable results on VQA. 

For images, we extract convolutional activation from the last layer of a 200-layer Residual Network~\cite{he2016deep}; for questions and answers, we extract the 300-dimensional \textsc{word2vec}~\cite{mikolov2013distributed} embedding for each words in a question/answer and compute their average as the feature. We then concatenate these features to be the input to the MLP model. Besides the Visual QA model that takes I, Q, and C as input, we also train two models that use only Q + C and C alone as the input. These two models can serve as $h_\cst{SD}$ described in Sect~\ref{S_method}.

Using simple models like MLP and average \textsc{word2vec} embeddings adds credibility to our studies---if models with limited capacity can latch on to the bias, models with higher capacity can only do better in memorizing the bias.

\vspace{-10pt}
\paragraph{Domain adaptation model}
We parameterize the transformation $g_q(\cdot)$, $g_a(\cdot)$ as a one hidden-layer MLP model (with 128 hidden nodes and ReLU) with residual connections directly from input to output. Such a design choice is due to the fact that the target embedding can already serve as a good starting point of the transforms.  We approximate  the $m(\textsf{TD}\rightarrow \textsf{SD})$ measure by adversarially learning a one hidden-layer MLP model (with 8,192 hidden nodes and ReLU) for binary classification between the source and the transformed target domain data, following the same architecture as the classifier in \emph{Name That Dataset!} game. 

For all our experiments on training $g_q(\cdot)$, $g_a(\cdot)$ and approximating $m(\textsf{TD}\rightarrow \textsf{SD})$, we use Adam~\cite{kingma2014adam} for stochastic gradient-based optimization. See the Suppl. for details.

\vspace{-10pt}
\paragraph{Domain adaptation settings}
As mentioned in Sect.~\ref{S_bias}, VQA (as well as VQA2) has around 30\% of the IQA triplets with the correct answers to be either ``Yes'' or ``NO''. On the other hand, Visual7W, COCOQA, and VG barely have triplets with such correct answers. Therefore, we remove those triplets from VQA and VQA2, leading to a reduced dataset VQA$^-$ and VQA2$^-$ that has 153,047/76,034 and 276,875/133,813 training/validation triplets, respectively.

We learn the Visual QA model using the training split of the source dataset and learn the domain adaptation transform using the training split of both datasets. 

\vspace{-10pt}
\paragraph{Other implementation details}
Questions in Visual7W, COCOQA, VG, VQA$^-$, and VQA2$^-$ are mostly started with the 6$W$ words. The frequencies, however, vary among datasets. To encourage $g_q$ to focus on matching the phrasing style rather than transforming one question type to the others, when training the binary classifier for $m(\textsf{TD}\rightarrow \textsf{SD})$ with Adams, we perform weighted sampling instead of uniform sampling from the source domain---the weights are determined by the ratio of frequency of each of the 6$W$ question types between the target and source domain. This trick makes our algorithm more stable.

%% file: exp_results.tex
\begin{table}[t]
\small
\tabcolsep 1.5pt
\centering
\caption{Domain adaptation (DA) results (in $\%$) on \emph{original} VQA ~\cite{agrawal2016vqa} and Visual7W~\cite{zhu2016visual7w}. \textbf{Direct}: direct transfer without DA. \cite{sun2016return}: CORAL. \cite{tzeng2017adversarial}: ADDA. \textbf{Within}: apply models trained on the target domain if supervised data is provided. (best DA result in bold)}

\begin{tabular}{|c|c|c|ccccc|c|}\hline
\multicolumn{9}{|c|}{VQA$^-$ $\rightarrow$ Visual7W} \\ \cline{1-9}
Direct & \cite{sun2016return} & \cite{tzeng2017adversarial} & [Q] & [T] & [T+D]  & [Q+T] & [Q+T+D] & Within \\
\hline
53.4 & 53.4 & 54.1 & 53.6& 54.5 & 55.7 & 55.2 & \textbf{58.5} & 65.7 \\         
\hline
\end{tabular}

\begin{tabular}{|c|c|c|ccccc|c|}\hline
 \multicolumn{9}{|c|}{Visual7W $\rightarrow$ VQA$^-$} \\ \cline{1-9}
Direct& \cite{sun2016return} & \cite{tzeng2017adversarial} & [Q] & [T] & [T+D]  & [Q+T] & [Q+T+D]& Within \\
\hline
 28.1 & 26.9 & 29.2 & 28.1 & 29.7 & 33.6 & 29.4 & \textbf{35.2} & 55.6 \\         
\hline
\end{tabular}
\vskip -5pt
\label{t_dataset_1}
\end{table}

\begin{table}[t]
	\small
	\tabcolsep 1.5pt
	\centering
	\caption{Domain adaptation (DA) results (in $\%$) on \emph{revised} VQA and Visual7W~\cite{chao2017being}. (best DA result in bold)}
	\begin{tabular}{|c|c|c|ccccc|c|}\hline
		\multicolumn{9}{|c|}{VQA$^-$ $\rightarrow$ Visual7W}  \\ \cline{1-9}
		Direct & \cite{sun2016return} & \cite{tzeng2017adversarial} & [Q] & [T] & [T+D]  & [Q+T] & [Q+T+D] & Within \\
		\hline
		46.1 & 47.2 & 47.8 & 46.2& 47.6 & 47.6 & 48.4 & \textbf{49.3} & 52.0  \\ 
		\hline
	\end{tabular}
	
	\begin{tabular}{|c|c|c|ccccc|c|}\hline
		\multicolumn{9}{|c|}{Visual7W $\rightarrow$ VQA$^-$} \\ \cline{1-9}
		Direct& \cite{sun2016return} & \cite{tzeng2017adversarial} & [Q] & [T] & [T+D]  & [Q+T] & [Q+T+D]& Within \\
		\hline
		45.6 & 45.3 & 45.9 & 45.9 & 45.9 & 47.8 & 45.8 & \textbf{48.1} & 53.7 \\ 
		\hline
	\end{tabular}
	
	\label{t_dataset_2}
	\vskip -15pt
\end{table}

We experiment on the five domain adaptation (DA) settings introduced in Sect.~\ref{DA_task} using the proposed algorithm. We also compare with ADDA~\cite{tzeng2017adversarial} and CORAL~\cite{sun2016return}, two DA algorithms that can also learn transformations from the target to the source domain and achieves comparable results on many benchmark datasets. Specifically, we learn two transformations to match the (joint) distribution of the questions and target answers. \emph{We only report the best performance among the five settings for ADDA and CORAL.} Table~\ref{t_dataset_1} and Table~\ref{t_dataset_2} summarize the results on the original and revised datasets, together with \textbf{Direct} transfer without any domain adaptation and \textbf{Within} domain performance where the Visual QA model is learned using the \emph{supervised data (i.e., IQA triplets)} of the target domain. Such supervised data is inaccessible in the adaptation settings we considered.

\vspace{-10pt}
\paragraph{Domain mismatch hurts cross-dataset generalization}
The significant performance drop in comparing \textbf{Within} domain and \textbf{Direct} transfer performance suggests that the learned Visual QA models indeed exploit certain domain-specific bias that may not exist in the other datasets. Such a drop is much severe between the original datasets than the revised datasets. Note that the two versions of datasets are different only in the decoys, and the revised datasets create decoys for both datasets by the same automatic procedure. Such an observation, together with the finding from \emph{Name That Dataset!} game, indicate that decoys contribute the most to the domain mismatch in Visual QA. 

\begin{table}[t]
	\small
	\tabcolsep 1.5pt
	\centering
	\caption{DA results (in $\%$) on \emph{revised} datasets, with target data sub-sampling by 1/16. FT: fine-tuning. (best DA result in bold)}
	
	\begin{tabular}{|c|c|c|ccccc|c|c|}
		\hline
		\multicolumn{10}{|c|}{VQA$^-$ $\rightarrow$ Visual7W}  \\ \cline{1-10}
		Direct & \cite{sun2016return} & \cite{tzeng2017adversarial} & [Q] & [T] & [T+D]  & [Q+T] & [Q+T+D] & Within & FT \\
		\hline
		46.1 & 45.6 & 47.8 & 46.1& 47.5& 47.6 & 48.3 & \textbf{49.1} & 39.7 & 48.3 \\         
		\hline
	\end{tabular}
	
	\begin{tabular}{|c|c|c|ccccc|c|c|}
		\hline
		\multicolumn{10}{|c|}{Visual7W $\rightarrow$ VQA$^-$} \\ \cline{1-10}
		Direct  & \cite{sun2016return} & \cite{tzeng2017adversarial} & [Q] & [T] & [T+D]  & [Q+T] & [Q+T+D] & Within & FT \\
		\hline
		45.6 & 44.8& 45.6 & 46.0 & 45.9 & 47.8 & 45.8 & \textbf{48.0} & 43.1 & 48.2 \\         
		\hline
	\end{tabular}
	\vskip -5pt
	\label{t_dataset_3}
\end{table}

\begin{table}[t]
	\small
	\tabcolsep 1.5pt
	\centering
	\caption{DA results (in $\%$) on on \emph{revised} datasets using a variant of the SMem~\cite{xu2016ask} model.}
	
	\begin{tabular}{|c|c|c|c|c|c|}
		\hline
		\multicolumn{3}{|c|}{VQA$^-$ $\rightarrow$ Visual7W} & \multicolumn{3}{|c|}{Visual7W $\rightarrow$  VQA$^-$ } \\ \cline{1-6}
		Direct & [Q+T+D] & Within & Direct & [Q+T+D] & Within \\
		\hline
		48.6 & 51.2 & 52.8 & 46.6 & 48.4 & 58.6\\       
		\hline
	\end{tabular}
	\vspace{-15pt}
	\label{t_attention}
\end{table}

\begin{table*}[t]
	\centering
	\small
	\tabcolsep 3pt
	\caption{Transfer results (in $\%$) across different datasets (the decoys are generated according to~\cite{chao2017being}). The setting for domain adaptation (DA) is on [Q+T+D] using \textbf{1/16} of the training examples of the target domain.}
	\label{t_across}
	\begin{tabular}{c|ccc|ccc|ccc|ccc|ccc}
		\cline{2-16}
		& \multicolumn{3}{c|}{Visual7W} & \multicolumn{3}{c|}{VQA$^-$} & \multicolumn{3}{c|}{VG} & \multicolumn{3}{c|}{COCOQA} & \multicolumn{3}{c}{VQA2$^-$} \\ \hline
		Training/Testing & Direct & DA & Within & Direct & DA & Within & Direct & DA & Within & Direct & DA & Within & Direct & DA & Within \\
		\hline
		Visual7W & 52.0 & - & - & 45.6& 48.0 & 43.1& 49.1& 49.4& 48.0& 58.0 & 63.1& 65.2& 43.9& 45.5& 43.6\\
		\hline
		VQA$^-$ & 46.1& 49.1 & 39.7 & 53.7& -& -& 44.8& 47.4& 48.0& 59.0 & 63.4& 65.2 & 50.7& 50.6& 43.6\\
		\hline
		VG & 58.1& 58.3& 39.7 & 52.6& 54.6& 43.1& 58.5& -& -& 65.5& 68.8& 65.2 & 50.1& 51.3& 43.6 \\
		\hline
		COCOQA & 30.1& 35.5& 39.7& 35.1& 40.4& 43.1& 29.1& 33.1 & 48.0& 75.8 & -& - & 33.3 & 37.5& 43.6\\
		\hline
		VQA2$^-$ & 48.8& 50.8& 39.7& 55.2& 55.3& 43.1& 47.3& 49.1& 48.0& 60.3& 64.9& 65.2& 53.8& - & -\\
		\hline
	\end{tabular}
	\vspace{-15pt}
	\label{t_transfer_across}
\end{table*}

\vspace{-10pt}
\paragraph{Comparison on domain adaptation algorithms}
Our domain adaptation algorithm outperforms \textbf{Direct} transfer in all the cases. On contrary, CORAL~\cite{sun2016return}, which aims to match the first and second order statistics between domains, fails in several cases, indicating that for domain adaptation in Visual QA, it is crucial to consider higher order statistics.

We also examine setting $\lambda$ in Eq.~(\ref{DA_obj}) to 0 for the [T] and [Q+T] settings\footnote{When $\lambda=0$, D has no effect (i.e., [Q+T+D] is equivalent to [Q+T]).} (essentially ADDA~\cite{tzeng2017adversarial} extended to multiple modalities), which leads to a drop of $\sim 1 $\%, demonstrating the effectiveness of leveraging the source domain for discriminative learning. See the Suppl. for more details.

\vspace{-10pt}
\paragraph{Different domain adaptation settings}
Among the five settings, we see that [T] generally gives larger improvement over \textbf{Direct} than [Q], suggesting that the domain mismatch in answers hinder more in cross-dataset generalization.

Extra information on top of [T] or [Q] generally benefits the domain adaptation performance, with [Q+T+D] giving the best performance. Note that different setting corresponds to different objectives in Eq.~(\ref{DA_obj}) for learning the transformations $g_q$ and $g_a$.  Comparing [T] to [T+D], we see that adding D helps take more advantage of exploiting the source domain's Visual QA knowledge, leading to a $g_a$ that better differentiates the correct answers from the decoys. On the other hand, adding T to [Q], or vice versa, helps constructing a better measure to match the feature distribution between domains. 

\vspace{-10pt}
\paragraph{Domain adaptation using a subset of data}
The domain adaptation results presented in Table~\ref{t_dataset_1} and ~\ref{t_dataset_2} are based on learning the transformations using all the training examples of the source and target domain. We further investigate the robustness of the proposed algorithm under a limited number of target examples. We present the results using only 1/16 of the them in Table~\ref{t_dataset_3}. The proposed algorithm can still learn the transformations well under such a scenario, with a slight drop in performance (i.e., $<0.5\%$). In contrast, learning Visual QA models with the same amount of limited target data (assuming the IQA triplets are accessible) from scratch leads to significant performance drop. We also include the results by fine-tuning, which is infeasible in any setting of Table~\ref{t_dataset} but can serve as an upper bound.

\vspace{-10pt}
\paragraph{Results on sophisticated Visual QA model} We further investigate a variant of the spatial memory network (SMem)~\cite{xu2016ask} for Visual QA, which utilizes the question to guide the visual attention on certain parts of the image for extracting better visual features. The results are shown in Table~\ref{t_attention}, where a similar trend of improvement is observed.

\vspace{-10pt}
\paragraph{Qualitative analysis}
The question type (out of the 6$W$ words) that improves the most from Direct to DA, when transferring from VQA$-$ to Visual7W in Table~\ref{t_dataset_1} using [Q+T+D], is ``When'' (from 41.8 to 63.4, while Within is 80.3). Other types improve $1.0\sim5.0$. This is because that the ``When''-type question is scarcely seen in VQA$-$, and our DA algorithm, together with the weighted sampling trick, significantly reduces the mismatch of question/answer phrasing of such a type. See the Suppl. for other results.

%% file: exp_across.tex

\label{S_transfer_across}
We perform a more comprehensive study on transferring the learned Visual QA models across five different datasets. We use the \emph{revised} candidate answers for all of them to reduce the mismatch on how the decoys are constructed. We consider the [Q+T+D] setting, and limit the disclosed target data to \textbf{1/16} of its training split size. The models for \textbf{Within} are also trained on such a size, using the supervised IQA triplets. Table~\ref{t_across} summarizes the results, where rows/columns correspond to the source/target domains.

On almost all (source, target) pairs, domain adaptation (DA) outperforms \textbf{Direct}, demonstrating the wide applicability and robustness of our algorithm. The exception is on (VQA$^-$, VQA2$^-$), where DA degrades by 0.1\%. This is likely due to the fact that these two datasets are constructed similarly and thus no performance gain can be achieved. Such a case can also be seen between Visual7W and VG. Specifically, domain adaptation is only capable in better transferring the knowledge learned in the source domain, but cannot acquire novel knowledge in the target domain.

The reduced training size significantly limits the performance of training from scratch (\textbf{Within}). In many cases \textbf{Within} is downplayed by DA, or even by \textbf{Direct}, showing the essential demand to leverage source domain knowledge. Among the five datasets, Visual QA models trained on VG seems to generalize the best---the DA results to any target domain outperforms the corresponding \textbf{Within}---indicating the good quality of VG.

In contrast, Visual QA models trained on COCOQA can hardly transfer to other datasets---none of its DA results to other datasets is higher than \textbf{Within}. It is also interesting to see that none of the DA results from other source domain (except VG) to COCOQA outperforms COCOQA's \textbf{Within}. This is, however, not surprising given how differently in the way COCOQA is constructed; i.e., the questions and answers are automatically generated from the captions in MSCOCO. Such a significant domain mismatch can also be witnessed from the gap between \textbf{Direct} and DA on any pair that involves COCOQA. The performance gain by DA over \textbf{Direct} is on average over 4.5\%, larger than the gain of any other pair, further demonstrating the effectiveness of our algorithms in reducing the mismatch between domains.

%% file: disc.tex
\section{Conclusion}
\label{S_disc}
We study cross-dataset adaptation for visual question answering. We first analyze the causes of bias in existing datasets. We then propose to reduce the bias via domain adaptation so as to improve cross-dataset knowledge transfer. To this end we propose a novel domain adaptation algorithm that minimizes the domain mismatch while leveraging the source domain's Visual QA knowledge. Through experiments on knowledge transfer among five popular datasets, we demonstrate the effectiveness of our algorithm, even under limited and fragment target domain information.

%% file: ack.tex
\footnotesize{ \paragraph{Acknowledgment} This work is partially supported by USC Graduate Fellowship, NSF IIS-1065243, 1451412, 1513966/1632803, 1208500, CCF-1139148, a Google Research Award, an Alfred. P. Sloan Research Fellowship and ARO\# W911NF-12-1-0241 and W911NF-15-1-0484. }

%% file: suppl_algorithms.tex
\section{\emph{Details on Name that Dataset!}}
\label{s_NTD}

As mentioned in Sect. 3.2 of the main text, we train a one-hidden-layer MLP to perform binary classification for detecting the origin of an input IQA triplet. The hidden layer is of 8,192 nodes and with the ReLU activation. The output of the MLP is normalized into $[0, 1]$ via the sigmoid function, and we train the MLP with the logistic (cross entropy) loss.
We use the penultimate layer of ResNet-200~\cite{he2016deep} as visual features to represent I and the average \textsc{word2vec} embeddings~\cite{mikolov2013distributed} as text features to represent Q and each C $\in$ A, as in~\cite{jabri2016revisiting}. We represent the whole set of decoys (denoted as D) in A by the average of those decoys' features\footnote{Visual7W~\cite{zhu2016visual7w} has 3 decoys per triplet and VQA~\cite{antol2015vqa} has 17 decoys. For fair comparison, we subsample 3 decoys for VQA. We then average the \textsc{word2vec} embedding of each decoy to be the feature of decoys.}. The input to the MLP is the concatenation of features from I, Q, T, and D (or a subset of them). The size of the training/validation/test triplets is 80,000/10,000/40,000, half from each dataset (i.e., either VQA~\cite{antol2015vqa} or Visual7W~\cite{zhu2016visual7w}).

\section{Details on the Proposed Domain Adaptation Algorithm}
\label{s_Alg}

\subsection{Approximating the JSD divergence}
As mentioned in Sect. 4.2 of the main text, we use the Jensen-Shannon Divergence (JSD) to measure the domain mismatch between two domains according to their empirical distributions. Dependent on the domain adaptation (DA) setting, the empirical distribution is computed on the (transformed) questions, (transformed) correct answers, or both.

Since JSD is hard to compute, we approximate it by training a binary classifier $\textsf{WhichDomain}(\cdot)$ to detect the domain of a question Q, a correct answer T, or a QT pair, following the idea of Generative Adversarial Network~\cite{goodfellow2014generative}. The architecture of $\textsf{WhichDomain}(\cdot)$ is exactly the same as that used for \emph{Name that dataset!}, except that the input features of examples from the target domain are after the transformations $g_q(\cdot)$ and $g_a(\cdot)$. 

\subsection{Details on the proposed algorithm}
We summarize the proposed domain adaptation algorithm for Visual QA under Setting[Q+T+D] in Algorithm~\ref{alg_1}. Algorithms of the other settings can be derived by removing the parts corresponding to the missing information.

\begin{algorithm*}
\label{alg_1}
\noindent\rule{16cm}{0.4pt}\\
\textbf{Notations} Denote the features of Q, T, D by $f_q$, $f_t$, and $f_d$. \emph{The D here stands for one decoy.}\\
\vspace{10pt}
\textbf{Goal} Learn transformations $g_q(\cdot)$, $g_a(\cdot)$ and a binary domain classifier $\textsf{WhichDomain}(\cdot)$, where $\vphi_q$, $\vphi_a$, and $\vtheta$ are the parameters to learn, respectively. $\textsf{WhichDomain}(\cdot)$ gives the conditional probability of being from the source domain\;	
\vspace{10pt}
 \For{number of training iterations}{
 Initialize the parameters $\vtheta$ of $\textsf{WhichDomain}(\cdot)$\;
 \For{k steps}{
 	Sample a mini-batch of $m$ pairs $\{Q_\cst{SD}^{(j)},T_\cst{SD}^{(j)}\}_{j=1}^m\sim\cst{SD}$\;
 	Sample a mini-batch of $m$ pairs $\{Q_\cst{TD}^{(j)},T_\cst{TD}^{(j)}\}_{j=1}^m\sim\cst{TD}$\;
 	Update $\textsf{WhichDomain}(\cdot)$ by ascending its stochastic gradient\;
 	$\nabla_{\vtheta}\left\{\frac{1}{m}\sum_{j=1}^m \Big[\log \textsf{WhichDomain}(\{{f_q}_\cst{SD}^{(j)},{f_t}_\cst{SD}^{(j)}\})\right.
 	+\left.\log(1-\textsf{WhichDomain}(\{g_q({f_q}_\cst{TD}^{(j)}),g_a({f_t}_\cst{TD}^{(j)})\}))\Big]\right\}$	
 }
 \For{l steps}{
 Sample a mini-batch of $m$ triplet $\{Q_\cst{TD}^{(j)},T_\cst{TD}^{(j)},D_\cst{TD}^{(j)}\}_{j=1}^m\sim\cst{TD}$\;
 Update the \textbf{transformations} by descending their stochastic gradients\;
	 $\nabla_{\vphi_q, \vphi_a}\left\{\frac{1}{m}\sum_{i=1}^m \log(1-\textsf{WhichDomain}(\{g_q({f_q}_\cst{TD}^{(j)}),g_a({f_t}_\cst{TD}^{(j)})\}))\right.+\left. \lambda \left( \ell \big(\{g_q({f_q}_\cst{TD}^{(j)}),g_a({f_t}_\cst{TD}^{(j)})\}) + \ell \big(\{g_q({f_q}_\cst{TD}^{(j)}),g_a({f_d}_\cst{TD}^{(j)})\}) \right) \right\}$
 }
 }
\vskip -5pt
 \noindent\rule{16cm}{0.4pt}
\vspace{5pt}
\caption{The proposed domain adaptation algorithm for Setting[Q+T+D]. $D_\cst{TD}^{(j)}$ denotes a single decoy. When the decoys of the target domain are not provided (i.e., Setting[Q+T]), the $\ell$ term related to $D_\cst{TD}^{(j)}$ is ignored.}
\end{algorithm*}

\section{Details on the Experimental Setup}
\label{s_set}
\subsection{Implementation details}
For all our experiments on training $g_q(\cdot)$, $g_a(\cdot)$, and $\textsf{WhichDomain}(\cdot)$, we use Adam~\cite{kingma2014adam} for stochastic gradient-based optimization, with learning rate = 10$^{-4}$ and mini-batch size = 100. We set $\lambda=0.5$ for Setting[Q+T+D] and Setting[T+D], and 0.1 for the others. We set $k=500$, and $l=5$, and train for 1,000 iterations.

\subsection{Domain adaptation settings}
Note that the ``Yes'' or ``NO'' issue we consider between VQA~\cite{antol2015vqa}, VQA2~\cite{goyal2016making} and Visaul7W~\cite{zhu2016visual7w}, Visual Genome (VG)~\cite{krishna2016vg}, COCOQA~\cite{ren2015exploring} is orthogonal to the one addressed in~\cite{goyal2016making,zhang2016yin}, which deal with the prior of answers within a single dataset.

\subsection{Sophisticated Visual QA models}
\label{s_sophi}
In the main text we experiment with a variant of the spatial memory network (SMem)~\cite{xu2016ask}. Instead of computing the visual attention for each word of the question, we directly compute the visual attention for the question using the average \textsc{word2vec} embeddings. We then concatenate the resulting visual features with the features of the question and a candidate answer (in the same way as the Visual QA model in Sect. 5.2 of the main text) as the input to train a one-hidden-layer MLP for binary classification.

We choose to train an MLP with candidate answers as a part of the input rather than training a multi-way classifier for the top frequent answers because the answer distributions can vary drastically across different domains  or datasets. In such a case, an IQA triplet of the target domain can never be answered correctly by the learnt multi-way classifier on the source domain if the correct answer is not in the top frequent answers of the source domain. 

In the Supplementary Material, we further experiment with a variant of the HieCoAtt model~\cite{lu2016hierarchical}, which applies the attention mechanism not only to images but also to questions (e.g., which word or phrase is more important). We extract the HieCoAtt features by removing the last layer (i.e., the multi-way classifier) of the HieCoAtt model, and concatenate the features again with the average \textsc{word2vec} embeddings of the question and a candidate answer to train an MLP for binary classification. The cross-dataset results are presented in Sect.~\ref{ss_HieCoAtt}. \emph{Note that we conduct this experiment not to achieve better performance, but to show that the dataset bias will also hinder cross-dataset generalization for more sophisticated models.}  

%% file: suppl_results.tex
\section{Additional Experimental Results}
\label{s_exp}

\subsection{The effect of the discriminative loss surrogate}
We provide in Table~\ref{t_lambda} the domain adaptation results on the [T] and [Q+T] settings when $\lambda$ is set to 0 (cf. Eq. (6) of the main text), which corresponds to omitting the discriminative loss surrogate $\hat{\ell}_\textsf{TD}$. In most of the cases, the results with $\lambda=0.1$ outperforms $\lambda=0$, showing the effectiveness of leveraging the source domain for
discriminative learning. Also note that when D is provided for the target domain (i.e., [T+D] or [Q+T+D]), it is the $\hat{\ell}_\textsf{TD}$ term that utilizes the information of D, leading to better results than [T] or [Q+T], respectively.

\begin{table}[t]
\small
\tabcolsep 2.5pt
\centering
\caption{Domain adaptation (DA) results (in $\%$) with or without the discriminative loss surrogate term}
\vspace{5pt}
\begin{tabular}{l|cc|cc}
\multicolumn{5}{c}{\textbf{original}} \\
\cline{2-5}
& \multicolumn{2}{|c}{VQA$^-$ $\rightarrow$ Visual7W} & \multicolumn{2}{|c}{Visual7W $\rightarrow$ VQA$^-$}  \\ \cline{1-5}
Setting  &[T] & [Q+T] &[T] & [Q+T]  \\
\hline
$\lambda=0$ & 54.1&  54.1& 29.2& 28.8\\
$\lambda=0.1$ & 54.5& 55.2 & 29.7& 29.4 \\         
\hline
\end{tabular}

\begin{tabular}{l|cc|cc}
\multicolumn{5}{c}{\textbf{revised}} \\
\cline{2-5}
& \multicolumn{2}{|c}{VQA$^-$ $\rightarrow$ Visual7W} & \multicolumn{2}{|c}{Visual7W $\rightarrow$ VQA$^-$}  \\ \cline{1-5}
Setting  &[T] & [Q+T] &[T] & [Q+T]  \\
\hline
$\lambda=0$ & 47.8&  47.8& 45.9& 45.7\\
$\lambda=0.1$ & 47.6& 48.4 & 45.9 & 45.8 \\         
\hline
\end{tabular}

\label{t_lambda}
\end{table}

\begin{figure}
	\centering
	\small
	\includegraphics[width=0.5\textwidth]{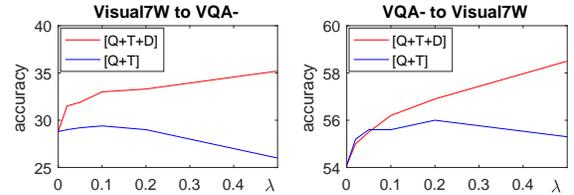}
	\caption{Results by varying $\lambda$ on the original VQA and Visual7W datasets, for both the [Q+T] and [Q+T+D] settings.}
	\label{f_lambda}
\end{figure}

We further experiment on different values of $\lambda$, as shown in Fig.~\ref{f_lambda}. For [Q+T], we achieve consistent improvement for $\lambda\leq0.1$. For [Q+T+D], we can get even better results by choosing a larger $\lambda$ (e.g. $\lambda=0.5$). 

\subsection{Domain adaptation using a subset of data}

Following Table 5 of the main text, we include in Table~\ref{t_dataset_3_supp} the results on the original VQA and Visual7W datasets, with target data sub-sampling by 1/16.

\begin{table}[t]
	\small
	\tabcolsep 1.5pt
	\centering
	\caption{DA results (in $\%$) on \emph{original} datasets, with target data sub-sampling by 1/16. FT: fine-tuning. (best DA result in bold)}
	\vspace{5pt}
	\begin{tabular}{|c|c|c|ccccc|c|c|}
		\hline
		\multicolumn{10}{|c|}{VQA$^-$ $\rightarrow$ Visual7W}  \\ \cline{1-10}
		Direct & \cite{sun2016return} & \cite{tzeng2017adversarial} & [Q] & [T] & [T+D]  & [Q+T] & [Q+T+D] & Within & FT \\
		\hline
		53.4 & 52.6 & 54.0 & 53.6& 54.4& 56.3 & 55.1 & \textbf{58.2} & 53.9  & 60.1 \\         
		\hline
	\end{tabular}
	
	\begin{tabular}{|c|c|c|ccccc|c|c|}
		\hline \multicolumn{10}{|c|}{Visual7W $\rightarrow$ VQA$^-$} \\ \cline{1-10}
		Direct & \cite{sun2016return} & \cite{tzeng2017adversarial} & [Q] & [T] & [T+D]  & [Q+T] & [Q+T+D] & Within & FT \\
		\hline
		28.1 & 26.5 & 28.8 & 28.1& 29.3 & 33.4 & 29.2 &\textbf{35.2} & 44.1 & 47.9\\         
		\hline
	\end{tabular}
	\vskip -10pt
	\label{t_dataset_3_supp}
\end{table}

\begin{figure*}[t]
\centering
\includegraphics[width=.45\textwidth]{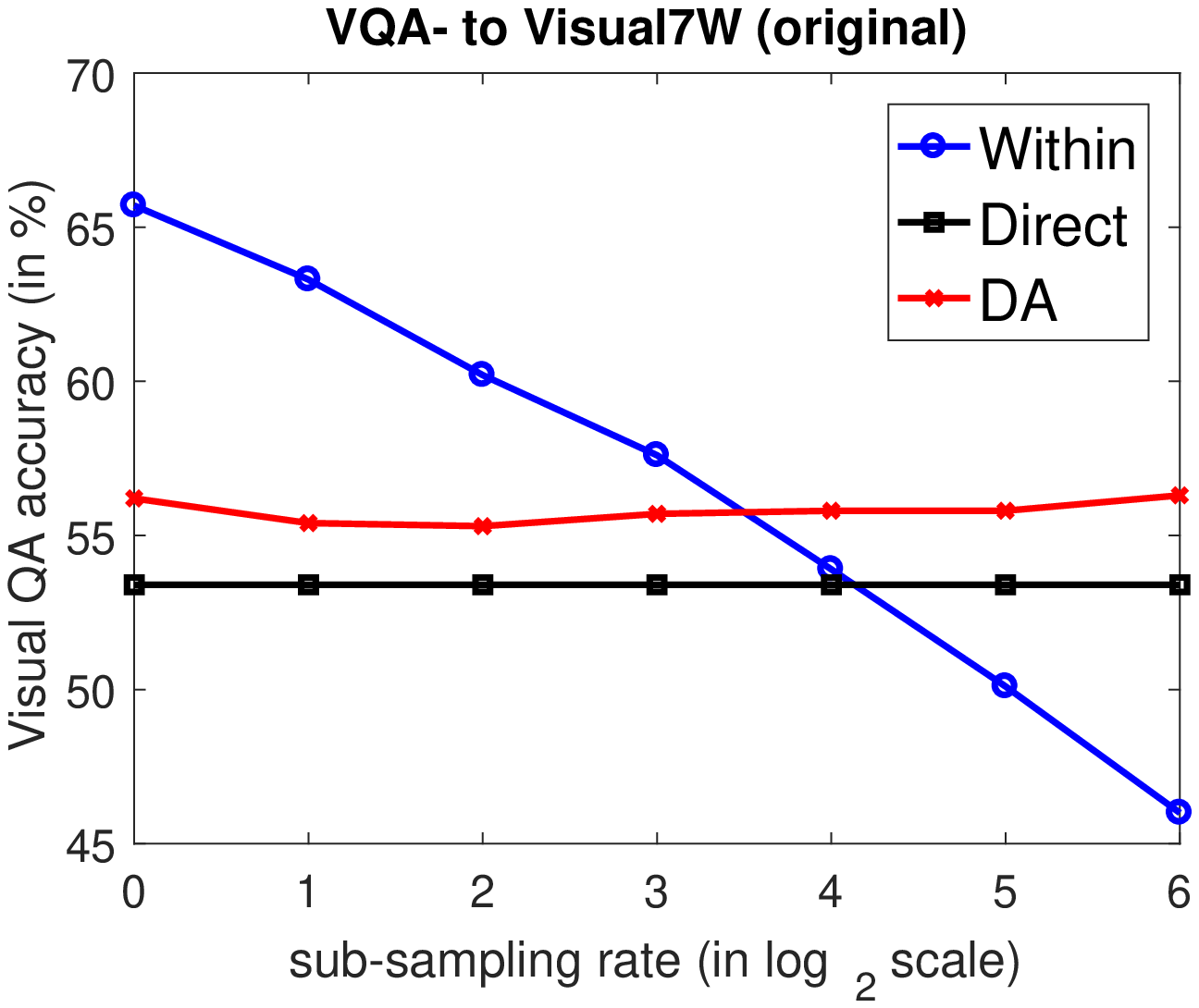}
\includegraphics[width=.45\textwidth]{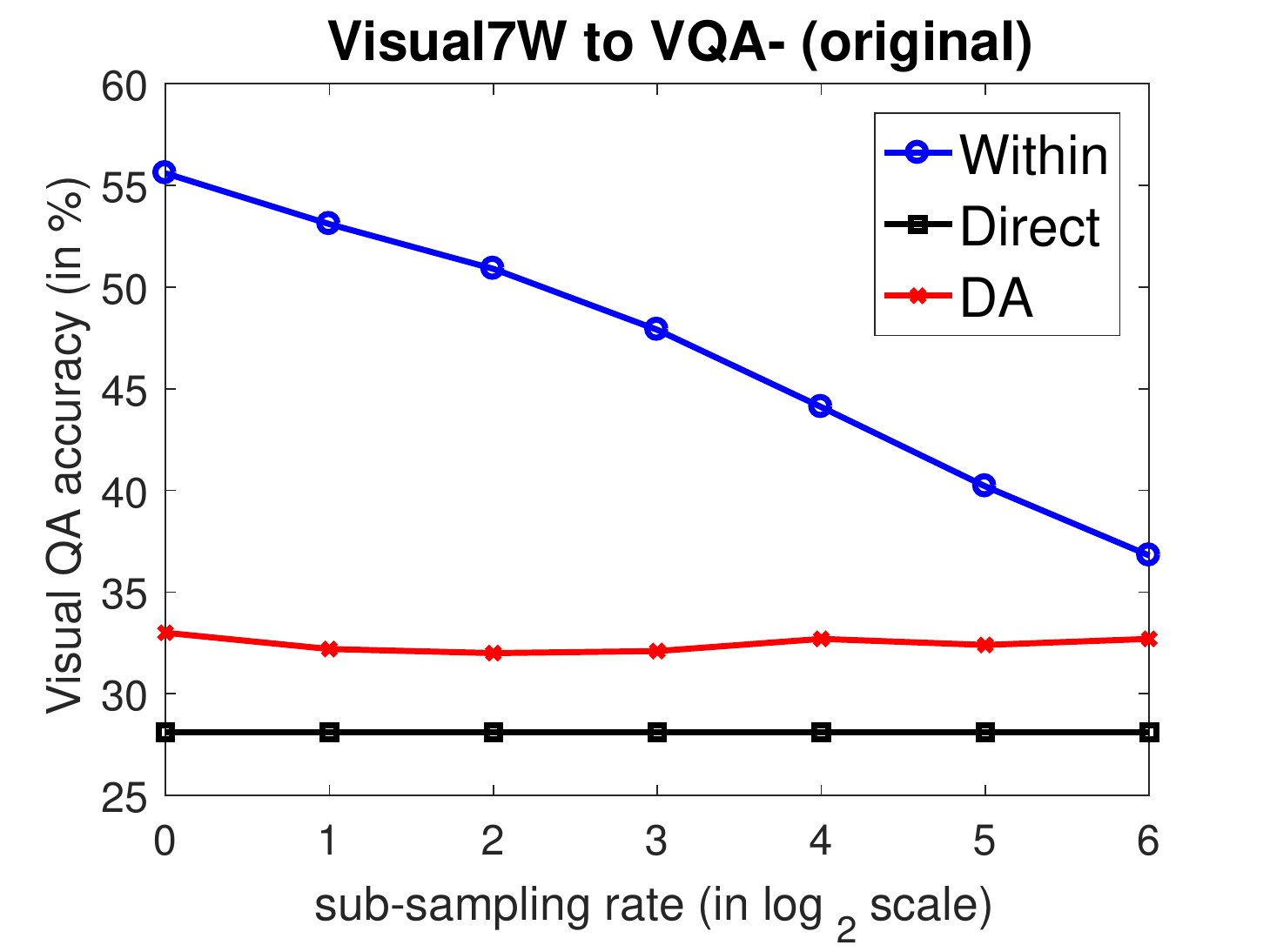}
\includegraphics[width=.45\textwidth]{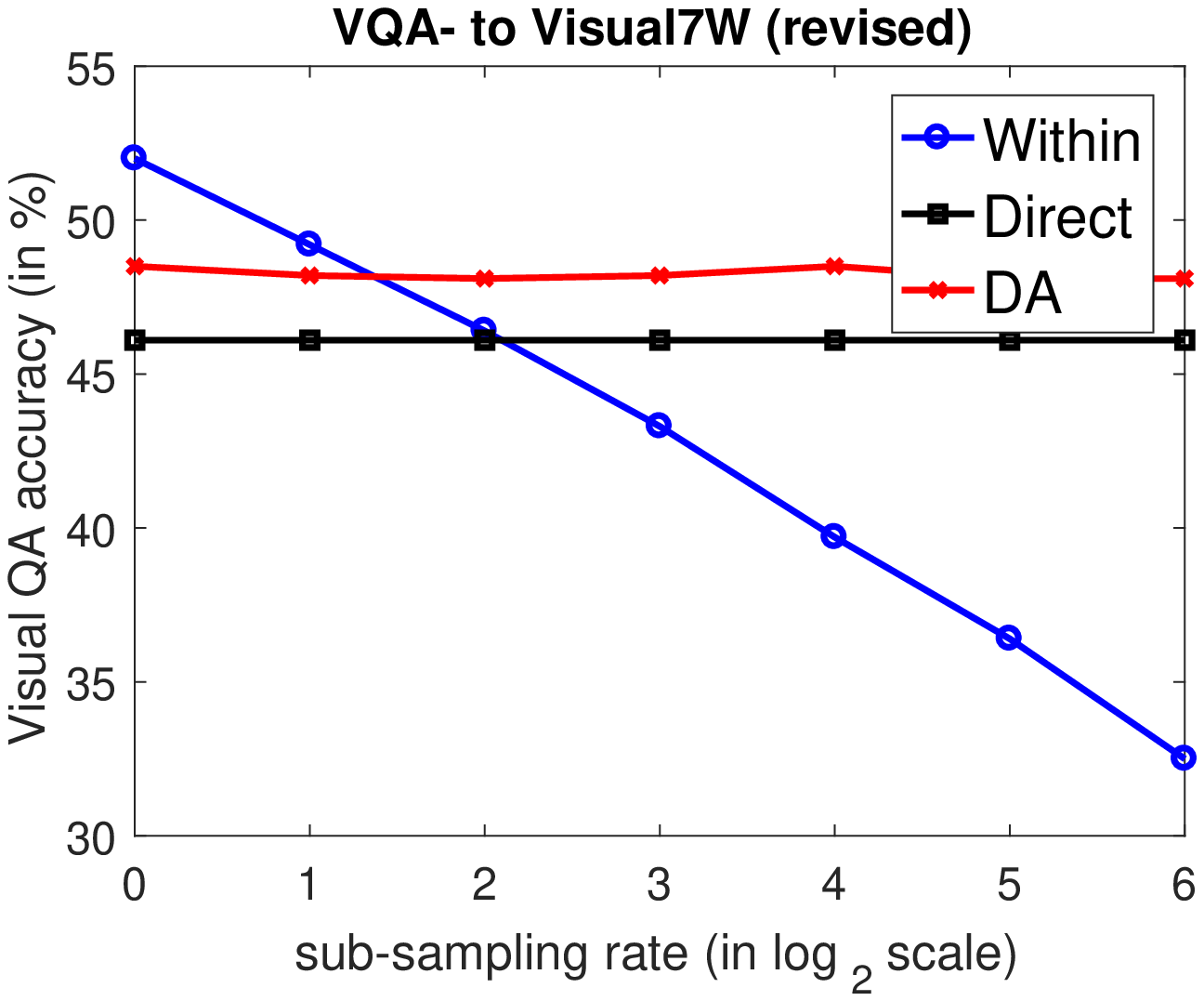}
\includegraphics[width=.45\textwidth]{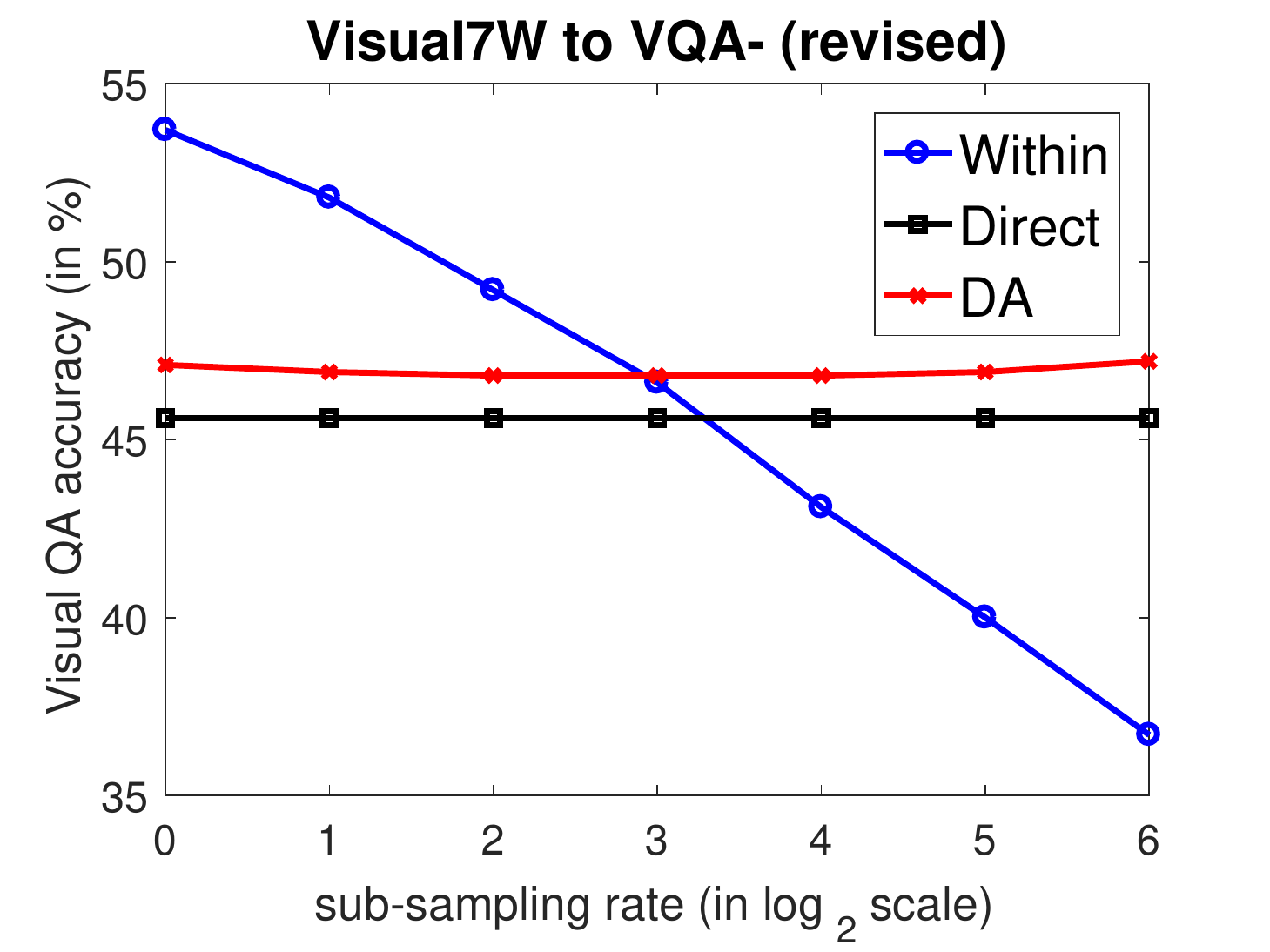}

\caption{\small Domain adaptation (DA) results (in $\%$) with limited target data, under Setting[Q+T+D] with $\lambda=0.1$. A sub-sampling rate $a$ means using $\cfrac{1}{2^a}$ of the target data.}

\label{fig:sub-sample}
\end{figure*}

We further consider domain adaptation (under Setting[Q+T+D] with $\lambda=0.1$) between Visual7W~\cite{zhu2016visual7w} and VQA$^-$~\cite{antol2015vqa} for both the original and revised decoys using $\cfrac{1}{2^a}$ of training data of the target domain, where $a\in[0,1,\cdots,6]$. The results are shown in Fig.~\ref{fig:sub-sample}. Note that the \textbf{Within} results are from models trained on the same sub-sampled size using the supervised IQA triplets from the target domain.

As shown, our domain adaptation (DA) algorithm is highly robust to the accessible data size from the target domain. On the other hand, the \textbf{Within} results from models training from scratch significantly degrade when the data size decreases. Except the case Visual7W $\rightarrow$ VQA$^-$ (original), domain adaptation (DA) using our algorithm outperforms the \textbf{Within} results after a certain sub-sampling rate. For example, on the case VQA$^-$ $\rightarrow$ Visual7W (revised), DA already outperforms \textbf{Within} under $\cfrac{1}{4}$ of the target data.

\subsection{Sharing transformations degrades the performance}
Although both questions and answers are text-based, they may have different degrees of domain mismatch (as shown in Table 1 of the main text). Ignoring such a fact and learning a single shared transformation degrades the performance. In Table 3 of the main text, the result on [Q+T+D] degrades from 56.2 to 55.6.

\subsection{Results on sophisticated Visual QA models}
\label{ss_HieCoAtt}

Following Table 6 of the main text, we include in Table~\ref{t_attention_supp} the results of SMem~\cite{xu2016ask} on the original datasets.

We further experiment with a variant of the HieCoAtt model~\cite{lu2016hierarchical} for Visual QA across datasets. See Sect.~\ref{s_sophi} for more details. The results are shown in
Table~\ref{t_hiecoatt}, where a similar trend of performance drop by \textbf{Direct} transfer and improvement by domain adaptation (in the [Q+T+D] setting) to those shown in the main text is observed.

\begin{table}[t]
	\small

	\tabcolsep 1.5pt
	\centering
	\caption{DA results (in $\%$) on on \emph{original} datasets using a variant of the SMem~\cite{xu2016ask} model.}
	\vspace{5pt}
		\begin{tabular}{|c|c|c|c|c|c|}

			\hline
			\multicolumn{3}{|c|}{VQA$^-$ $\rightarrow$ Visual7W} & \multicolumn{3}{|c|}{Visual7W $\rightarrow$  VQA$^-$ } \\ \cline{1-6}
			Direct & [Q+T+D] & Within & Direct & [Q+T+D] & Within\\
			\hline
			56.3 & 61.0 & 65.9& 27.5& 34.1 & 58.5\\     
			\hline
		\end{tabular}

	\label{t_attention_supp}
\end{table}

\begin{table}[t]
\small
\tabcolsep 1.5pt
\centering
\caption{DA results (in $\%$) on VQA and Visual7W (both original and revised) using a variant of the HieCoAtt model~\cite{lu2016hierarchical}.}
\vspace{5pt}

\begin{tabular}{|c|c|c|c|c|c|}
\multicolumn{6}{c}{\textbf{original}} \\
\hline
\multicolumn{3}{|c|}{VQA$^-$ $\rightarrow$ Visual7W} & \multicolumn{3}{|c|}{Visual7W $\rightarrow$  VQA$^-$ } \\ \cline{1-6}
Direct & [Q+T+D] & Within & Direct & [Q+T+D] & Within\\
\hline
 51.5& 56.2 & 63.9& 27.2& 33.1& 54.8\\     
\hline
\end{tabular}

\begin{tabular}{|c|c|c|c|c|c|}

\multicolumn{6}{c}{\textbf{revised}} \\
\hline
\multicolumn{3}{|c|}{VQA$^-$ $\rightarrow$ Visual7W} & \multicolumn{3}{|c|}{Visual7W $\rightarrow$  VQA$^-$ } \\ \cline{1-6}
Direct & [Q+T+D] & Within & Direct & [Q+T+D] & Within \\
\hline
 46.4& 48.2 & 51.5 & 44.5& 46.3& 55.6\\        
\hline
\end{tabular}

\label{t_hiecoatt}
\end{table}

\subsection{Open-ended (OE) results}

We apply Visual QA models learned with the multiple-choice setting to evaluate on the open-ended one (i.e., select an answer from the top frequent ones, or from the set of all possible answers in the training data). The result on transferring from VQA$^-$ to COCOQA is in Table~\ref{t_OE}. Our adaptation algorithm still helps transferring.

\begin{table}[t]
	\small

	\tabcolsep 1pt
	\centering
	\caption{\small OE results (VQA$^-$ $\rightarrow$ COCOQA, sub-sampled by 1/16).}
	\vspace{5pt}
	\begin{tabular}{|c|c|c|}
		\hline

		Direct & [Q+T+D] & Within \\
		\hline
		16.7 & 24.0 & 26.9\\
		\hline
	\end{tabular}
	\vspace{-10pt}
	\label{t_OE}
\end{table}

\subsection{Cross-dataset results across five datasets}
Table~\ref{st_transfer_across} summarizes the results of the same study as in Sect. 5.4 of the main text, except that now \textbf{all} the training examples of the target domain are used. The models for \textbf{Within} are also trained on such a size, using the supervised IQA triplets.

Compared to Table 7 of the main text, we see that the performance drop of DA from using all the training examples of the target domain to $1/16$ of them is very small (mostly smaller than $0.3\%$), demonstrating the robustness of our algorithm under limited training data. On the other hand, the drop of \textbf{Within} is much more significant---for most of the (source, target) pairs, the drop is at least $10\%$.

\begin{table*}[t]
	\centering
	\small
	\tabcolsep 3pt
	\caption{Transfer results (in $\%$) across datasets. The decoys are generated according to~\cite{chao2017being}, where each IQT triplet is accompanied by 6 decoys (the accuracy of random guess is $14.3\%$). The setting for domain adaptation (DA) is on [Q+T+D] using \textbf{all} the training examples of the target domain.}
	\vspace{5pt}
	\begin{tabular}{c|ccc|ccc|ccc|ccc|ccc}
		\cline{2-16}
		& \multicolumn{3}{c|}{Visaul7W~\cite{zhu2016visual7w}} & \multicolumn{3}{c|}{VQA$^-$~\cite{antol2015vqa}} & \multicolumn{3}{c|}{VG~\cite{krishna2016vg}} & \multicolumn{3}{c|}{COCOQA~\cite{ren2015exploring}} & \multicolumn{3}{c}{VQA2$^-$~\cite{goyal2016making}} \\ \hline
		Training/Testing & Direct & DA & Within & Direct & DA & Within & Direct & DA & Within & Direct & DA & Within & Direct & DA & Within \\
		\hline
		Visual7W~\cite{zhu2016visual7w} & 52.0 & - & - & 45.6 & 48.1 & 53.7& 49.1& 49.6& 58.5& 58.0 & 63.0& 75.8& 43.9& 45.6& 53.8\\
		\hline
		VQA$^-$~\cite{antol2015vqa} & 46.1& 49.3& 52.0 & 53.7& -& -& 44.8& 47.9& 58.5& 59.0 & 64.7& 75.8 & 50.7& 50.6& 53.8\\
		\hline
		VG~\cite{krishna2016vg} & 58.1& 58.4& 52.0 & 52.6& 54.4& 53.7& 58.5& -& -& 65.5& 68.8& 75.8 & 50.1& 51.5& 53.8 \\
		\hline
		COCOQA~\cite{ren2015exploring} & 30.1& 34.4& 52.0& 35.1& 40.2& 53.7& 29.1& 33.4 & 58.5& 75.8 & -& - & 33.3 & 37.9& 53.8\\
		\hline
		VQA2$^-$~\cite{goyal2016making} & 48.8& 51.0& 52.0& 55.2& 55.3& 53.7& 47.3& 49.6& 58.5& 60.3& 65.2& 75.8& 53.8& - & -\\
		\hline
	\end{tabular}
	\label{st_transfer_across}
\end{table*}

For most of the (source, target) pairs shown in Table~\ref{st_transfer_across}, \textbf{Within} outperforms \textbf{Direct} and DA. The notable exceptions are (VG, Visual7W) and (VQA2$^-$, VQA$^-$). This is likely due to the fact that VG and Visual7W are constructed similarly while VG has more training examples than Visual7W. The same fact applies to VQA2$^-$ and VQA$^-$. Therefore, the Visual QA model learned on the source domain can be directly applied to the target domain and leads to better results than \textbf{Within}. 

\subsection{Qualitative results}
\begin{figure}
	\centering
	\small
	\includegraphics[width=0.5\textwidth]{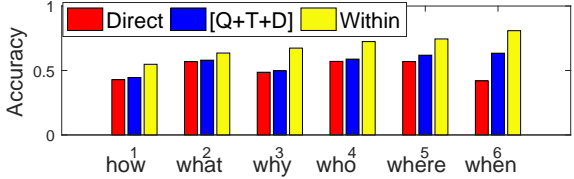}
	\caption{Qualitative comparison on different type of questions, following the analysis of Sect. 5.3 of the main text when transferring from VQA$-$ to Visual7W (on the original datasets).}
	\label{f_qual}
\end{figure}

Following the analysis of Sect. 5.3 of the main text, we shown in Fig~\ref{f_qual} the results on each question type when transferring from VQA$-$ to Visual7W (on the original datasets). [Q+T+D] outperforms \textbf{Direct} at all the question types.

